\documentclass[11pt,a4paper]{article}
\usepackage[hyperref]{acl2020}
\usepackage{times}
\usepackage{latexsym}

\usepackage{url}
\usepackage{graphicx}
\usepackage{amsmath}
\usepackage{amsfonts}
\usepackage{amssymb}
\usepackage{bbm}
\usepackage{booktabs}
\usepackage{xcolor}
\usepackage{enumitem}
\usepackage{siunitx}

\newcommand{\minus}{\scalebox{0.75}[1.0]{$-$}}

\aclfinalcopy 

\title{Evaluating Explainable AI: Which Algorithmic Explanations \\ Help Users Predict Model Behavior?}

\author{Peter Hase \and Mohit Bansal \\
  UNC Chapel Hill \\
  \texttt{peter@cs.unc.edu, mbansal@cs.unc.edu}}

\begin{document}
\maketitle
\begin{abstract}

Algorithmic approaches to interpreting machine learning models have proliferated in recent years. We carry out human subject tests that are the first of their kind to isolate the effect of algorithmic explanations on a key aspect of model interpretability, \emph{simulatability}, while avoiding important confounding experimental factors. A model is simulatable when a person can predict its behavior on new inputs. Through two kinds of simulation tests involving text and tabular data, we evaluate five explanations methods: (1) LIME, (2) Anchor, (3) Decision Boundary, (4) a Prototype model, and (5) a Composite approach that combines explanations from each method. Clear evidence of method effectiveness is found in very few cases: LIME improves simulatability in tabular classification, and our Prototype method is effective in counterfactual simulation tests. We also collect subjective ratings of explanations, but we do not find that ratings are predictive of how helpful explanations are. 
Our results provide the first reliable and comprehensive estimates of how explanations influence simulatability across a variety of explanation methods and data domains. We show that (1) we need to be careful about the metrics we use to evaluate explanation methods, and (2) there is significant room for improvement in current methods.\footnote{We make all our supporting code, data, and models publicly available at: {\scriptsize \url{https://github.com/peterbhase/InterpretableNLP-ACL2020}}}

\end{abstract}

\section{Introduction}

Interpretable machine learning is now a widely discussed topic \cite{rudin_stop_2019, doshi-velez_towards_2017, lipton_mythos_2016, gilpin_explaining_2018}. While survey papers have not converged on definitions of ``explainable" or ``interpretable," there are some common threads in the discourse. Commentators observe that interpretability is useful for achieving other model desiderata, which may include building user trust, identifying the influence of certain variables, understanding how a model will behave on given inputs, and ensuring that models are fair and unbiased.

In their review, \citet{doshi-velez_towards_2017} outline an approach to measuring interpretability. They describe two human-subject tasks that test for a particularly useful property: \emph{simulatability}. A model is simulatable when a person can predict its behavior on new inputs. This property is especially useful since it indicates that a person understands \emph{why} a model produces the outputs it does. The first of the two tasks is termed \emph{forward simulation}: given an input and an ``explanation," users must predict what a model would output for the given input. The second is \emph{counterfactual simulation}: users are given an input, a model's output for that input, and an ``explanation" of that output, and then they must predict what the model will output when given a perturbation of the original input. The explanation itself is algorithmically generated by a method for interpreting or explaining a model. Simulation tests have been carried out before, but no study to date has isolated the effect of explanations on simulatability \cite{ribeiro_anchors_2018, chandrasekaran_explanations_2018, nguyen_comparing_2018, bang_explaining_2019}. 

\begin{figure*}[t]
    \centering
    \includegraphics[width=0.99\textwidth]{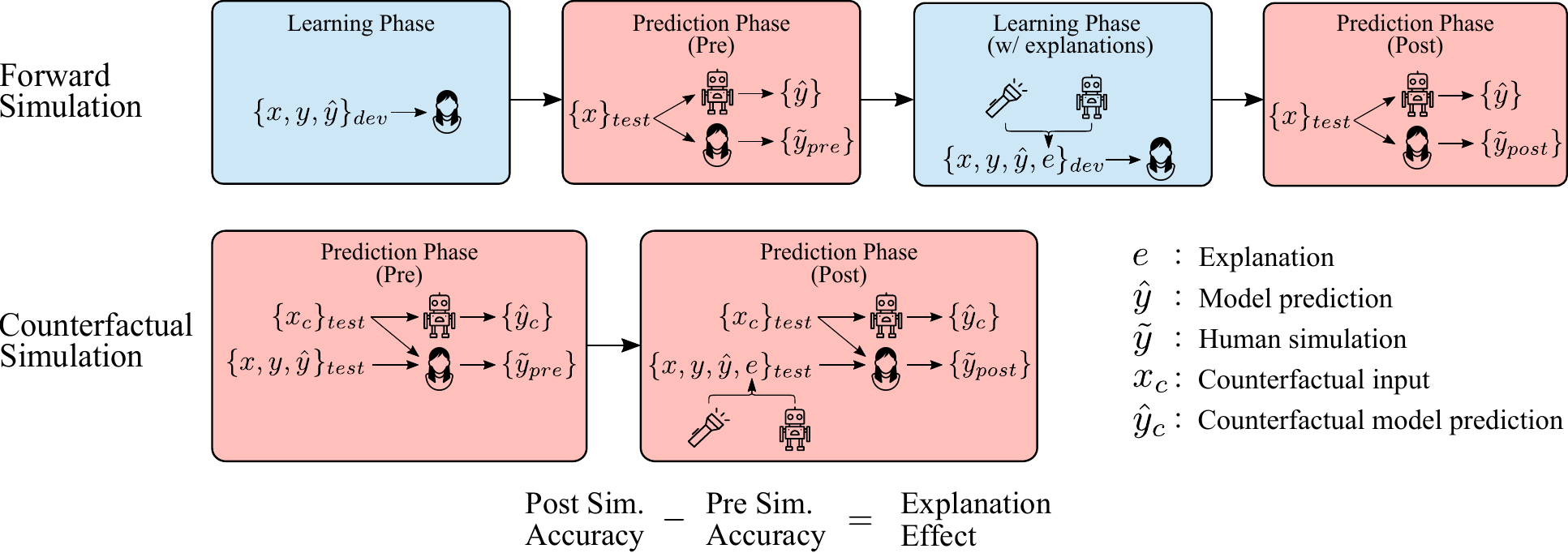}
    \caption{Forward and counterfactual simulation test procedures. We measure human users' ability to predict model behavior. We isolate the effect of explanations by first measuring baseline accuracy, then measuring accuracy after users are given access to explanations of model behavior. In the forward test, the explained examples are distinct from the test instances. In the counterfactual test, each test instance is a counterfactual version of a model input, and the explanations pertain to the original inputs.}
    \label{fig:procedure}
    \vspace{-8pt}
\end{figure*}

We carry out simulation tests that are the first to incorporate all of the following design choices: (1) separating explained instances from test instances, so explanations do not give away the answers, (2) evaluating the effect of explanations against a baseline of \emph{unexplained} examples, (3) balancing data by model correctness, so users cannot succeed by guessing the true label, and (4) forcing user predictions on all inputs, so performance is not biased toward overly specific explanations. We display our study design in Figure \ref{fig:procedure}.

We provide results from high-quality human user tests (with over 2100 responses) that include both forward and counterfactual simulation tasks. Through these tests, we measure explanation effectiveness for five methods across text and tabular classification tasks. 
Our evaluation includes two existing explanation techniques, LIME and Anchor \cite{ribeiro_why_2016, ribeiro_anchors_2018}, and we translate two other explanation methods from image recognition models to work with our textual and tabular setups. The first of these is a latent space traversal method, which we term the Decision Boundary approach \cite{joshi_xgems:_2018, samangouei_explaingan:_2018}, and the second is a case-based reasoning method, which we term the Prototype method \cite{chen_this_2018}. The final method is a novel Composite approach that combines complementary explanations from each method. Lastly, we also collect subjective, numerical user ratings of explanation quality. Our key findings are:
\vspace{0pt}
\begin{enumerate}[nosep, wide=0pt, leftmargin=*, after=\strut]
    \item LIME improves forward and counterfactual simulatability in our tabular classification task.
    \item Prototype improves counterfactual simulatability across textual and tabular data domains.
    \item No method definitively improves forward and counterfactual simulatability together on the text task, though our Prototype and Composite methods perform the best on average.
    \item It appears that users' quality ratings of explanations are not predictive of how helpful the explanations are with counterfactual simulation.
    \item While users rate Composite explanations as among the best in quality, these combined explanations do not overtly improve simulatability in either data domain.
\end{enumerate}

\section{Background and Related Work}

\subsection{What Does ``Interpretable" Mean?}

Survey papers use key terms in varying ways. \citet{rudin_stop_2019} draws a distinction between interpretability and explainability, suggesting that a model is interpretable if it performs computations that are directly understandable. Post-hoc explanations, on the other hand, are potentially misleading approximations of the true computations. \citet{gilpin_explaining_2018} also distinguish between the two concepts, though they define them differently.

In this paper, we do not distinguish between interpretability and explainability. Rather, we adopt the conceptual framework of \citet{doshi-velez_towards_2017}, who consider interpretability in terms of downstream desiderata one can assess models with respect to. Our terminology is as follows: we will say that \emph{explanation methods} may improve the \emph{interpretability} of a model, in the sense that an interpretable model is \emph{simulatable}.

\subsection{Explanation Methods}

Several taxonomies have been proposed for categorizing methods for interpretability. We organize methods below into the categories of: feature importance estimation, case-based reasoning, and latent space traversal. 

\vspace{2pt}
\noindent\textbf{Feature Importance Estimation.} \ Feature importance estimates provide information about how the model uses certain features. Most prominent among these methods are the gradient-based approaches first introduced for vision by \citet{simonyan_deep_2013}, which \citet{li_visualizing_2016} show may be translated for use with text data. These approaches have since been demonstrated to sometimes behave in counterintuitive ways \cite{adebayo_sanity_2018, kim_interpretability_2018}. 
A number of alternative methods have been proposed for quantifying feature importance across data domains \cite{kim_interpretability_2018, lundberg_unified_2017, sundararajan_axiomatic_2017}. In our study, we choose to evaluate two domain-agnostic approaches, LIME and Anchor \cite{ribeiro_why_2016, ribeiro_anchors_2018}. These methods use simple models, i.e. sparse linear models and rule lists, to approximate complex model behavior locally around inputs. They show the estimated effects of directly interpretable features on the model's output. For these methods, what is ``local" to an input is defined in a domain-specific manner via a perturbation distribution centered on that input.

\vspace{2pt}
\noindent\textbf{Case-based Reasoning.} \ Prototype models classify new instances based on their similarity to other known cases. Two works on prototype models for computer vision introduced neural models that learn prototypes corresponding to parts of images \cite{chen_this_2018, hase_interpretable_2019}. These prototypes are used to produce classifier features that are intended to be directly interpretable.

\vspace{2pt}
\noindent\textbf{Latent Space Traversal.} \  These methods traverse the latent space of a model in order to show how the model behaves as its input changes. In a classification setting, crossing the decision boundary may reveal necessary conditions for a model's prediction for the original input. Several methods exist for vision models \cite{joshi_xgems:_2018, samangouei_explaingan:_2018}. To our knowledge no such approach exists for discriminative models of text and tabular data, so we develop a simple method for these kinds of models (described in Section \ref{sec:db}).

\subsection{Evaluating Interpretability}

Here we discuss works involving automatic and human evaluations of interpretability, as well as how we improve on past simulation test design. While human evaluations are useful for evaluating many aspects of interpretability, we restrict our discussion to works measuring simulatability.

\vspace{2pt}
\noindent\textbf{Improving Forward Test Design.}  \ Forward simulation tasks have been implemented in many different forms, and there is a serious need for consensus on proper procedure here. \citet{doshi-velez_towards_2017} originally propose that users predict model behavior, given an input and an explanation. With many explanation methods, this is a trivial task \textit{because the explanations directly reveal the output}. For example, LIME gives a predicted probability that indicates the model behavior with high likelihood.
We make a number of experimental design choices that give us more reliable estimates of method effectiveness than past studies. (1) We separate the explained instances from the test instances, to prevent explanations from giving away the answers. In three studies, the same data points were used as both explanation and prediction items \cite{nguyen_comparing_2018, chandrasekaran_explanations_2018, bang_explaining_2019}. (2) We evaluate the effect of explanations against a baseline where users see the same example data points without explanations. No prior evaluation includes this control. (3) Two choices further distinguish our test from that of \citet{ribeiro_anchors_2018}. We balance data by model correctness, so users cannot succeed simply by guessing the true label, and we force user predictions on every input, so our metrics do not favor overly niche explanations.

\vspace{2pt}
\noindent\textbf{Counterfactual Simulatability.} \ Counterfactual simulatability has, to our knowledge, never been measured for machine learning models. While \citet{doshi-velez_towards_2017} propose asking users to edit inputs in order to change the model outputs, we instead ask users to predict model behavior on edited versions of data points, as this approach is more scalable than soliciting creative responses. 

\vspace{2pt}
\noindent\textbf{Relation to Automatic Tests.} \ Prior works have proposed automatic metrics for feature importance estimates \cite{nguyen_comparing_2018, hooker_benchmark_2019, deyoung_eraser_2019}. Typically these operate by checking that model behavior follows reasonable patterns on counterfactual inputs constructed using the explanation, e.g., by masking ``important" features and checking that a class score drops. Whereas automatic metrics define appropriate model behavior in advance for counterfactual instances generated by a fixed schema, we present a random counterfactual to a human and elicit their prediction of model behavior for that instance. This allows for human validation of model behavior in a broader range of input scenarios than an automatic procedure, where human expectations are given in response to diverse and concrete examples rather than dictated in advance.

\begin{figure*}[th!]
\centering
 \includegraphics[width=.93\textwidth]{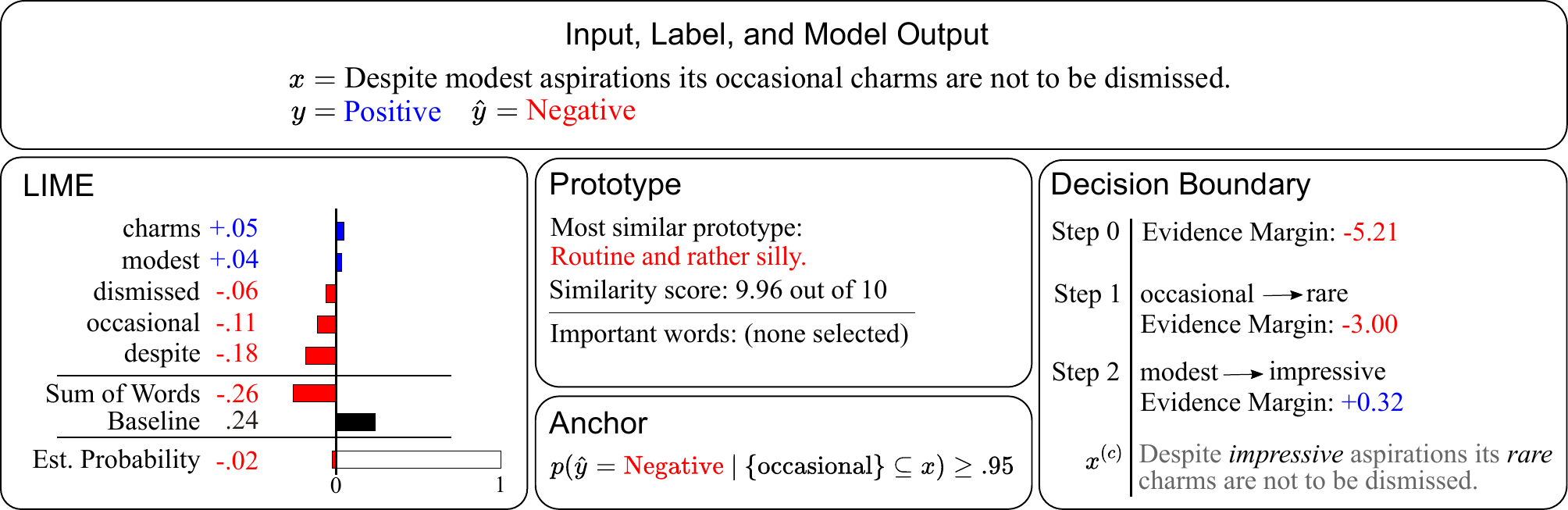}
 \vspace{-3pt}
\caption{Explanation methods applied to an input from the test set of movie reviews.} 
\label{fig:explanations}
\vspace{-8pt}
\end{figure*}

\vspace{2pt}
\noindent\textbf{Subjective Ratings.} \ \citet{hutton_crowdsourcing_nodate} measure user judgments of whether word importance measures explain model behavior in a text classification setting. Our rating task is thus similar to theirs; our changes are that we evaluate with a Likert scale rather than forced ranking, using explanation techniques for neural models rather than word importance estimates from a naive Bayes classifier. In another study, users judged image classification explanations on a Likert scale ranging from ``no explanation" to ``concise explanation" \cite{bang_explaining_2019}. Whereas this scale focuses on conciseness, we ask users to rate how explanations reveal reasons for model behavior.

\section{Explanation Methods}

In this section, we describe the explanation methods. Example explanations for a test movie review are shown in Figure \ref{fig:explanations}. We limit our discussion of LIME and Anchor, since details for these methods can be found in the original papers. Note that LIME, Anchor, and our Decision Boundary method can be used with arbitrary blackbox models. The Prototype method is itself a neural model that also produces an explanation.

\subsection{LIME}

\citet{ribeiro_why_2016} present LIME as a local linear approximation of model behavior. With a user-specified feature space, a linear model is fit to the blackbox outputs on samples from a distribution around an input. We set the number of features to use to 5, and we take class probabilities as our model output. When showing LIME explanations to users, we give them the selected features with estimated weights, the model intercept, the sum of model weights, and the predicted model output.

\subsection{Anchor}
\citet{ribeiro_anchors_2018} introduce a method for learning rule lists that predict model behavior with high confidence. With samples from a distribution around an input, they use a PAC learning approach to obtain a rule list. When the rules apply to an input, there is a high probability it will receive the same prediction as the original. The feature space of the rule list is specified by the user. As in the original work, we use individual tokens for our text data, and we use the same learning parameters for each Anchor explanation.

\subsection{Prototype Model}
Prototype models have previously been used for interpretable computer vision \cite{chen_this_2018, hase_interpretable_2019}. We develop a prototype model for use with text and tabular classification tasks. In our model, a neural network $g$ maps inputs to a latent space, and the score of class $c$ is: 
\begin{align*}
\vspace{-15pt}
    f(\mathbf{x}_i)_c = \max_{\mathbf{p_k} \in P_c} a(g(\mathbf{x}_i), \mathbf{p_k})
    \vspace{-12pt}
\end{align*}
where $a$ is a similarity function for vectors in the latent space, and $P_c$ is the set of protoype vectors for class $c$. We choose the Gaussian kernel for our similarity function: $ a(\mathbf{z}_i, \mathbf{p_k}) = e^{-||\mathbf{z}_i - \mathbf{p_k}||^2}$. The model predicts inputs to belong to the same class as the prototype they're closest to in the latent space. Unlike in \citet{chen_this_2018}, we take the max activation to obtain concise explanations.

In lieu of image heatmaps, we provide feature importance scores. What distinguishes these scores from those of standard feature importance estimates is that the scores are prototype-specific, rather than class-specific. We choose a feature omission approach for estimation. With text data, omission is straightforward: for a given token, we take the difference in function output between the original input and the input with that token's embedding zeroed out. In the tabular domain, however, variables can never take on meaningless values. To circumvent this problem, we take the difference between the function value at the original input and the \emph{expected} function value with a particular feature missing. The expectation is computed with a distribution over possible values for a missing feature, which is provided by a multinomial logistic regression conditioned on the remaining covariates. 

When presenting prototype explanations, we provide users with the predicted class score, most similar prototype, and top six feature importance scores, provided that score magnitudes meet a small threshold. In the explanation in Figure \ref{fig:explanations}, no scores meet this threshold. We set the size of $P_c$ to 40 for our text classification task and 20 for our tabular classification task. For further training and feature importance details, see the Appendix.

\subsection{Decision Boundary}
\label{sec:db}

\citet{joshi_xgems:_2018} and \citet{samangouei_explaingan:_2018} introduce techniques for traversing the latent spaces of generative image models. Their methods provide paths that start at input data points and cross a classifier's decision boundary. Such methods may help users see the necessary conditions for the model prediction.

We provide a simple method for traversing the latent space of a discriminative classifier (see example in Figure \ref{fig:explanations}). Our algorithm first samples around the original input to get instances that cross the decision boundary. A counterfactual input is chosen from these by taking the instance with the fewest edited features (tokens or variables), while breaking ties using the Euclidean distance between latent representations. Lastly, we provide a path between inputs by greedily picking the edit from the remaining edits that least changes the model's evidence margin, which is the difference between positive and negative class scores. The explanations we present to users include the input, steps to the counterfactual input, and evidence margin at each step. When the path is longer than four steps, we show only the last four.

\subsection{Composite Approach}

We hypothesize that the above explanations provide complementary information, since they take distinct approaches to explaining model behavior. Hence, we test a Composite method that combines LIME and Anchor with our decision boundary and prototype explanations. We make two adjustments to methods as we combine them. First, we show only the last step of each decision boundary explanation, i.e., the set of changes that flips the prediction. Second, we train our prototype model with its feature extraction layers initialized from the neural task model and thereafter fixed. We do so since we are interested in explaining the task model behavior, and this tactic yields prototypes that reflect characteristics of the task model.

\begin{table*}[th]
    \small
    \centering
\begin{tabular*}{\textwidth}{l @{\extracolsep{\fill}} c cccSc c cccSc} 
\toprule
& & \multicolumn{5}{c}{Text} &  & \multicolumn{5}{c}{Tabular} \\
\cmidrule(lr){3-7} \cmidrule(lr){9-13}
Method & & $n$ & Pre & Change & CI & $p$ & & $n$ & Pre & Change & CI & $p$ \\
\midrule
\addlinespace
User Avg. & & 1144 & 62.67 & - & 7.07 & - & & 1022 & 70.74 & {-} & 6.96 & -  \\
\addlinespace
\midrule
LIME & & 190 & - & 0.99 & 9.58 & .834 & & 179 & - & \textbf{11.25} $\;$ & 8.83 & .014 \\
Anchor & & 181 & - & 1.71 & 9.43 & .704 & & 215 & - & 5.01 & 8.58 & .234 \\
Prototype & & 223 & - & 3.68 & 9.67 & .421 & & 192 & - & 1.68 & 10.07 & .711 \\
DB & & 230 & - & \minus 1.93 $\>$ & 13.25 & .756 & & 182 & - & 5.27 & 10.08 & .271\\
Composite & & 320 & - & 3.80 & 11.09 & .486 & & 254 & - & 0.33 & 10.30 & .952 \\    
\bottomrule
 \end{tabular*}
 \vspace{-6pt}
\caption{Change in user accuracies after being given explanations of model behavior, relative to the baseline performance (Pre). Data is grouped by domain. CI gives the 95\% confidence interval, calculated by bootstrap using $n$ user responses, and we bold results that are significant at a level of $p<.05$. LIME improves simulatability with tabular data. Other methods do not definitively improve simulatability in either domain.} 
\label{table:test_by_domain}

\end{table*}

\begin{table*}[th]
\begin{small}
    \centering
\begin{tabular*}{\textwidth}{l  @{\extracolsep{\fill}} c cccSc c cccSc} 
\toprule
& & \multicolumn{5}{c}{Forward Simulation} &  & \multicolumn{5}{c}{Counterfactual Simulation} \\
\cmidrule(lr){3-7} \cmidrule(lr){9-13}
Method & & $n$ & Pre & Change & CI & $p$ & & $n$ & Pre & Change & CI & $p$ \\
\midrule
\addlinespace
User Avg. & & 1103 & 69.71 & - & 6.16 & - & & 1063 & 63.13 & - & 7.87 & -  \\
\addlinespace
\midrule
LIME & & 190 & - & 5.70 & 9.05 & .197 & &  179 & - & 5.25 & 10.59 & .309 \\
Anchor & & 199 & - & 0.86 & 10.48 & .869 & & 197 & - & 5.66 & 7.91 & .140 \\
Prototype & & 223 & - & \minus 2.64 $\;$ & 9.59 & .566 & & 192 & - & \textbf{9.53} & 8.55  & .032 \\
DB & & 205 & - & \minus 0.92 $\;$ & 11.87 & .876 & & 207 & - & 2.48 & 11.62 & .667 \\
Composite & & 286 & - & \minus 2.07 $\;$ & 8.51 & .618 & & 288 & - & 7.36 & 9.38  & .122 \\    
\bottomrule
 \end{tabular*}
 \vspace{-6pt}
\caption{Change in user accuracies after being given explanations of model behavior, relative to the baseline performance (Pre). Data is grouped by simulation test type. CI gives the 95\% confidence interval, calculated by bootstrap using $n$ user responses. We bold results that are significant at the $p<.05$ level. Prototype explanations improve counterfactual simulatability, while other methods do not definitively improve simulatability for one test.}
\label{table:test_by_type}
\vspace{-8pt}
\end{small}
\end{table*}

\section{Experimental Design}

In this section, we describe our datasets, task models, user pool, and experimental design.

\subsection{Data and Task Models}

We perform experiments for classification tasks with text and tabular data. The first dataset consists of movie review excerpts \citep{pang_thumbs_2002}. The dataset includes 10,662 reviews with binary sentiment labels, which we split into partitions of 70\%, 10\%, and 20\% for the train, validation, and test sets, respectively. We use the same neural architecture as in \citet{yang_hierarchical_2016}, limited to use with single sentences. The second dataset is the tabular \textit{Adult} data from the UCI ML repository \cite{Dua:2019}. This dataset contains records of 15,682 individuals, and the label is whether their annual income is more than \$50,000. We use the same data processing scheme and neural network architecture as \citet{ribeiro_anchors_2018}. Model accuracies are given in the Appendix.

\subsection{User Pool}

We gathered over 2100 responses via in-person tests with 32 trained undergraduates who had taken at least one course in computer science or statistics.\footnote{We require this advanced background because explanations rely on conditional probabilities, approximations of probabilities, and other quantitative concepts.} Each user was randomly assigned to one of the ten conditions corresponding to our dataset-method pairs. Once each condition had at least 3 full tests collected, we allocated remaining participants to the Composite method. In order to ensure high quality data, we employed a screening test to check for user understanding of their explanation method and test procedure. Two participants were screened out due to low scores. We also excluded data from a user whose task completion time was extremely low. We paid all users \$15 USD per hour.
Ten users were tested again with a new dataset and explanation method, giving us a total of 39 user tests. Some users had to exit the experiment before finishing all of the tasks; for data analysis purposes, we consider only task items answered in both Pre and Post test phases.

\subsection{Simulation Tests}

We collect $1103$ forward test and $1063$ counterfactual test responses in total.

\begin{table*}[th]
\begin{small}
    \centering
\begin{tabular*}{\textwidth}{l @{\extracolsep{\fill}} c cSSS c cSSS} 
\toprule
& & \multicolumn{4}{c}{Text Ratings} & & \multicolumn{4}{c}{Tabular Ratings} \\
\cmidrule(lr){3-6} \cmidrule(lr){8-11}
Method & & $n$ & $\mu$ & CI &  $\sigma$ & & $n$ & $\mu$ & CI & $\sigma$ \\
\midrule
    LIME & & 144 & 4.78 & 1.47 & 1.76 & & 130 & 5.36 & .63 & 1.70 \\
    Anchor & & 133 & 3.86 & .59 & 1.79 & & 175 & 4.99 & .71 & 1.38  \\
    Prototype & & 191 & 4.45 & 1.02 & 2.08 & & 144 & 4.20 & .82 & 1.88 \\
    DB & & 224 & 3.85 & .60 & 1.81 & & 144 & 4.61 & 1.14 & 1.86 \\ 
    Composite & & 240 & 4.47 & .58 & 1.70 & & 192 & 5.10 & 1.04 & 1.42 \\
\bottomrule

 \end{tabular*}
 \vspace{-4pt}
\caption{User simulatability ratings by data domain, on a scale of 1 to 7. The mean and standard deviation for ratings are given by $\mu$ and $\sigma$. The 95\% confidence interval for the mean is given by CI, as calculated by bootstrap.} 
\label{table:ratings}
\vspace{-8pt}
\end{small}
\end{table*}

\vspace{1pt}
\noindent\textbf{Forward Simulation.} \ This test is represented in Figure \ref{fig:procedure}. The test is split into four phases: a learning phase, a Pre prediction phase, a learning phase \emph{with explanations}, and a Post prediction phase. To begin, users are given 16 examples from the validation set with labels and model predictions but no explanations. Then they must predict the model output for either 16 or 32 new inputs, with the number chosen based on user time constraints. Users are not allowed to reference the learning data while in prediction phases. Next, they return to the same learning examples, now with explanations included. Finally, they predict model behavior again on the same instances from the first prediction round. By design, any improvement in user performance in the Post prediction phase is attributable only to the addition of explanations. We show a screenshot of the user testing interface in the Appendix.

\vspace{1pt}
\noindent\textbf{Counterfactual Simulation.} \ Represented in Figure \ref{fig:procedure}, this test requires users to predict how a model will behave on a \emph{perturbation} of a given data point. The test consists of Pre and Post prediction rounds, where the only difference between them is the addition of explanations. In both rounds, we provide users with the same 32 inputs from the test dataset (or 16 due to time constraints), their ground truth labels, the model's prediction, and a perturbation of the input. See the Appendix for a description of the perturbation generation algorithm. Users then predict model behavior on the perturbations. In the Post round, users are given the same data, but they are also equipped with explanations of the model predictions for the original inputs. Therefore, any improvement in performance is attributable to the addition of explanations.

\vspace{1pt}
\noindent\textbf{Data Balancing.} \ One critical aspect of our experimental design is our data balancing. We aim to prevent users from succeeding on our tests simply by guessing the true label for every instance. To do so, we ensure that true positives, false positives, true negatives, and false negatives are equally represented in the inputs. Likewise, for the counterfactual test, we sample perturbations such that for any instance, there is a 50\% chance that the perturbation receives the same prediction as the original input. We confirm user understanding of the data balancing in our screening test.

\vspace{2pt}
\noindent\textbf{Data Matching.} \ Within each data domain, all users receive the same data points throughout the experiment. This design controls for any differences in the data across conditions and users, though this does reduce the information added by each test, making our confidence intervals relatively wide given the same sample size. We also match data across prediction rounds in order to control for the influence of particular data points on user accuracy between the Pre and Post phases.

\subsection{Subjective Simulatability Ratings}

Users see explanations in two phases of the tests: the second learning phase in the forward test, and the Post phase of the counterfactual test. In these stages, we ask users to give subjective judgments of the explanations. They rate each method on a 7 point Likert scale, in response to the question, ``Does this explanation show me why the system thought what it did?" We explain that users should give higher ratings when the explanation shows the reasons for a model prediction, regardless of whether or not the prediction is correct.

\section{Results}

We report data from a total of 2166 responses from 39 user tests. Each test is for a method and data domain pair, and contains either 16 or 32 task items, with some missingness due to users exiting the study early. In the results to follow, we use the term Change to refer to our estimate of explanation effectiveness: the difference in user accuracy across prediction phases in simulation tests. We perform two-sided hypothesis tests for this quantity by a block bootstrap, resampling both users and unique task items within each condition \cite{efron1994introduction}. In addition, since users complete the first prediction round in either simulation test without access to explanations, we estimate the mean Pre accuracy for each method with a random effects model. This allows us to share information across methods to yield more precise estimates of test performance. 

Below, we analyze our experimental results and answer three questions: 1) Do explanations help users? 2) How do users rate explanations? 3) Can users predict explanation effectiveness?

\subsection{Do explanations help users?}

We show simulation test results in Tables \ref{table:test_by_domain} and \ref{table:test_by_type}. In Table \ref{table:test_by_domain}, we group results by data domain, and in Table \ref{table:test_by_type}, we group results by test type. 

Our principal findings are as follows: 
\begin{enumerate}[nosep, wide=0pt, leftmargin=*, after=\strut]
    \item LIME with tabular data is the only setting where there is definitive improvement in forward and counterfactual simulatability. With no other method and data domain do we find a definitive improvement across tests.
    \item Even with combined explanations in the Composite method, we do not observe definitive effects on model simulatability. 
    \item Interestingly, our prototype method does reliably well on counterfactual simulation tests in both data domains, though not forward tests. It may be that the explanations are helpful only when shown side by side with inputs.
\end{enumerate}
These results suggest that: (1) many explanation methods may not noticeably help users understand how models will behave, (2) methods that are successful in one domain might not work equally well in another, (3) combining information from explanations does not result in overt improvements in simulatability. Yet, given our wide confidence intervals, these results should be considered cautiously. It may also be that other methods do in fact improve simulatability, but we have not precisely estimated this. For example, our Prototype and Composite methods do the best on average with text data, though we cannot be confident that they improve simulatability.

Note that estimates of explanation effectiveness could be influenced by users simply regressing to the mean accuracy between prediction rounds. We find that our primary results are not skewed by this phenomenon: the highest estimates of Change in each data domain and test type come from conditions where mean Pre test performance was either above the overall mean or, in one case, within 1.15\ percentage points. This potential problem is further mitigated by our random effects model of Pre test performance, which pulls low Pre test means toward the overall mean. 

\subsection{How do users rate explanations?}

It seems that, as intended, users rated explanations based on quality rather than model correctness, as we observe no significant difference in ratings grouped by model correctness (table in Appendix). In Table \ref{table:ratings}, we show user ratings for each method and data domain.

We observe that: 1) ratings are generally higher for tabular data, relative to text data, 2) the Composite and LIME methods receive the highest ratings in both domains, and 3) variance in explanation ratings is quite high, relative to their scale.

\subsection{Can users predict explanation effectiveness?}

We answer this question by measuring how explanation ratings relate to user correctness in the Post phase of the counterfactual simulation test. In this phase, users rate explanations of model predictions for an original input and predict model behavior for a perturbation of that input. If ratings of explanation quality are a good indicator of their effectiveness, we would expect to see that higher ratings are associated with user correctness.

We do not find evidence that explanation ratings are predictive of user correctness. We estimate the relationship via logistic regression with user correctness and ratings. We test models with both absolute ratings and ratings normalized within users, since ratings lack an absolute scale between users. With 640 text data points, we estimate with 95\% confidence that moving from a rating of 4 to 5 is associated with between a $\minus2.9$ and $5.2$ percentage point change in expected user correctness. Using normalized ratings, we find that moving from the mean explanation rating to the first standard deviation is associated with between a $\minus3.9$ and $12.2$ percentage point change. With 515 tabular data points, we estimate that a change in rating from 4 to 5 is associated with between a $\minus2.6$ and $5.3$ percentage point change in expected user correctness. Of course, we have not shown that there is no association. Yet it's important to note that if there is no relationship between user ratings and simulatability, then simply querying humans about explanation quality will not provide a good indication of true explanation effectiveness. 

\section{Qualitative Analysis}

When do explanations succeed at improving user accuracy, and when do they fail at doing so? Below, we present example counterfactual test items, and we analyze how the explanations may have pointed to the reasons for model behavior. 

\subsection{Explanation Success Example}

For the example below, 5 of 6 Post test responses for Prototype and LIME were correct that the model output did not change for the counterfactual, up from 3 of 6 in the Pre test.

\smallskip

\noindent \emph{Original} ($\hat{y} = pos$): ``Pretty much sucks, but has a funny moment or two." 

\smallskip

\noindent \emph{Counterfactual} ($\hat{y}_c = pos$): ``\emph{Mostly} \emph{just} \emph{bothers}, but \emph{looks} a funny moment or two."

\smallskip

\noindent \textbf{LIME} identifies ``funny" and ``moment" as positive words, with weights adding to $1.04$ after including the baseline. The notable negative word is ``sucks" ($w=-.23$), which changes to a similar word (``bothers"). All together, LIME suggests the prediction would stay the same since the positive words are unaffected and the only important negative word has a similar substitute.

The \textbf{Prototype} model gives the most activated prototype: ``\emph{Murders by Numbers} isn't a great movie, but it's a perfectly acceptable widget." It identifies ``but" and ``funny" as important words for the prototype's activation. The counterfactual is still similar to the prototype in key ways, suggesting the prediction would not change.

\subsection{Explanation Failure Example}

For the item below, only 7 of 13 responses were correct after seeing explanations, with no method improving correctness relative to the Pre test accuracy. Users needed to predict that the model prediction changed to negative for the counterfactual.

\smallskip

\noindent \emph{Original} ($\hat{y} = pos$): ``A bittersweet film, simple in form but rich with human events." 

\smallskip

\noindent \emph{Counterfactual} ($\hat{y}_c = neg$): ``A \emph{teary} film, simple in form but \emph{vibrant} with \emph{devoid} events."

\smallskip

\noindent \textbf{Anchor} gives one word as a condition for the original positive prediction: ``bittersweet." But what happens when ``bittersweet" changes to ``teary"? The Anchor explanation does not actually apply to this counterfactual scenario, as its probabilistic description of model behavior is conditioned on the word bittersweet being present. 

\textbf{LIME} gives five words, each with small weights ($|w|<.04$), while the baseline is $.91$. This suggests that LIME has failed to identify features of the input that are necessary to the model output. Among these five words are the three that changed between sentences, but we would not suspect from their weights that the changes made in the counterfactual would flip the model output.

\textbf{Decision Boundary} gives a counterfactual input with a negative prediction: ``A \emph{sappy} film, simple in \emph{link} but \emph{unique} with human events." However, it is difficult to tell whether this counterfactual sentence is similar in decision-relevant ways to the proposed counterfactual sentence. 

The \textbf{Prototype} model gives the activated prototype for the original prediction: ``Watstein handily directs and edits around his screenplay's sappier elements...and sustains \emph{Off the Hook}'s buildup with remarkable assuredness for a first-timer." No important words are selected. We are left without a clear sense of why this was the most similar prototype and what circumstances would lead to the model output changing. 

These examples reveal areas for improvement in explanations. Better methods will need to distinguish between sufficient and necessary factors in model behavior and clearly point to the ways in which examples share decision-relevant characteristics with new inputs. Further, they must do so in the appropriate feature space for the problem at hand, especially for models of complex data.

\section{Discussion}

\vspace{2pt}
\noindent\textbf{Forward Tests Stretch User Memory.} \ We show users 16 examples during learning phases but do not allow them to reference the learning data during prediction phases. Reasonably, some users reported that it was difficult to retain insights from the learning phase during later prediction rounds. 

\vspace{2pt}
\noindent\textbf{Generating Counterfactual Inputs.} \ It may be difficult to algorithmically construct counterfactual inputs that match the true data distribution, especially when seeking to change the model prediction. Our text counterfactuals are regularly out of the data distribution, in the sense that no real movie review would exhibit the word choice they do. We still consider these inputs to be of interest, for the reason that a model will handle such inputs in \emph{some} manner, and we aim to assess all possible model behaviors in our analysis.

\vspace{2pt}
\noindent\textbf{Fair Comparison of Explanation Methods.} \ In our forward simulation treatment phases, we provide users with 16 explained instances and allow them to read at their own pace. We control for the number of data points between methods, but one could instead control for user exposure time or computation time of explanation generation. 
Further, for LIME and Anchor, there are approaches for efficiently covering the space of inputs with a limited budget of examples \cite{ribeiro_anchors_2018}. We opt not to use them since 1) they are not applicable to the Decision Boundary and Prototype methods, which lack a similar notion of coverage, and 2) it is not clear whether these approaches are useful for text data. It may be that when using such approaches, LIME and Anchor perform better on forward simulation tasks.

\section{Conclusion}

Simulatability metrics give a quantitative measure of interpretability, capturing the intuition that explanations should improve a person's understanding of why a model produces its outputs. In this paper, we evaluated five explanation methods through simulation tests with text and tabular data. These are the first experiments to fully isolate the effect of algorithmic explanations on simulatability. We find clear improvements in simulatability only with LIME for tabular data and our Prototype method in counterfactual tests. It also appears that subjective user ratings of explanation quality are not predictive of explanation effectiveness in simulation tests. These results suggest that we must be careful about the metrics we use to evaluate explanation methods, and that there is significant room for improvement in current methods.

\vspace{-3pt}
\section*{Acknowledgments} 
\vspace{-3pt}
We thank the reviewers for their helpful feedback and our study users. This work was supported by NSF-CAREER Award 1846185, DARPA MCS Grant N66001-19-2-4031, a Royster Society PhD Fellowship, and Google and AWS cloud compute awards. The views contained in this article are those of the authors and not of the funding agency.

\bibliography{acl2020}
\bibliographystyle{acl_natbib}

\appendix

\section{Appendix}

\subsection{Method Implementations}

\paragraph{Explanation methods.} For our tabular data, we use the implementations of Anchor and LIME provided in the code for \citet{ribeiro_anchors_2018}. We implement our prototype and decision boundary methods. With text data, we use the implementation of Anchor provided by \citet{ribeiro_anchors_2018}, and for LIME we use the code provided with \citet{ribeiro_why_2016}. As before, we implement our prototype and decision boundary methods.

\paragraph{Text and Tabular Models.} We train neural networks for both tasks as follows: for our tabular task model, we use a neural network with two hidden layers, each of width 50, as \citet{ribeiro_anchors_2018} do. For our text task model, we use a BiLSTM of the kind introduced by \citet{yang_hierarchical_2016}, who reported state of the art results on a number of sentiment analysis tasks. Since their network is designed for classification of documents, we limit our network components to those relevant to classification of single sentences. We build our prototype models on top of the feature extractor layers of each of these models, meaning that we only replace the final classifier layer of the neural task model with a prototype layer. Accuracies for each model are shown in Table \ref{table:accuracy}. The task models are trained with stochastic gradient descent and a cross-entropy loss function, using early stopping on a validation dataset and $l_2$ regularization with a coefficient of $1\mathrm{e}\minus4$. See training details for the prototype models below. 

\paragraph{Prototype Model Training.} Here we describe our prototype training algorithm, beginning with weight initialization. We initialize 1) feature extraction layers using the pretrained weights of our neural task model, 2) prototype vectors via k-means clustering on the latent representations of the entire training set, and 3) final classifier weights as 1 where the corresponding prototype's class matches the weight vector's class, and $\minus.5$ elsewhere. The objective function for our prototype models contains three terms: 1) a cross entropy loss, 2) $l_1$ regularization on off-class weights in the classifier, and 3) a separation cost term, which is the minimum distance between a latent representation and any prototype not belonging to the input's class.

\begin{table}[t]
\centering
\begin{small}
\begin{tabular}{l c S}
  \toprule
  \multicolumn{3}{c}{Model Accuracies} \\
  \cmidrule(lr){1-3}
  {Data \& Model} & & {Test Acc}  \\
  \midrule
     Text \\
     \quad Task Model & & 80.93 \\
     \quad Prototype & & 80.64 \\
    \addlinespace[1pt]
    \midrule
    \addlinespace[2pt]
    Tabular \\
     \quad Task Model & & 83.49 \\
     \quad Prototype & & 81.90 \\
  \bottomrule
\end{tabular}
\caption{Model accuracies on each data domain. Text data is split into partitions of 70\%, 10\%, and 20\% for the train, validation, and test sets, respectively. We use the same data processing scheme as \citet{ribeiro_anchors_2018} for tabular data.}
\label{table:accuracy}
\end{small}
\end{table}

\begin{table}[t!]
\begin{small}
\centering
\begin{tabular}{l l l l l}
  \toprule
  & \multicolumn{4}{c}{User Ratings} \\
  \cmidrule(lr){2-5}
  Model Correctness & n & $\mu$ & CI & $\sigma$ \\
  \midrule
    Text \\
     \quad Correct & 464 & 4.44 & .49 & 1.89 \\
     \quad Incorrect & 468 & 4.12 & .67 & 1.81 \\
    \addlinespace[1pt]
        \midrule
    \addlinespace[2pt]
    Tabular \\
     \quad Correct & 391 & 5.09 & .27 & 1.64 \\
     \quad Incorrect & 394 & 4.64 & .27 & 1.69 \\
  \bottomrule
\end{tabular}
\caption{User simulatability ratings grouped by model correctness and data domain. Users do not seem to be rating explanations simply based on model correctness, as the differences in group means based on model correctness are not significant at a level of $p < .05$.}
\label{table:ratings_correctness}
\end{small}
\end{table}

\paragraph{Importance Scores in Protoype Model.} For a given feature, we compute an importance score by taking the difference in function output with that feature present in the input, relative to when that feature is omitted. With text data, there are a number of mechanisms by which one can omit a word from an input; we opt for setting that word's embedding to the zero vector. For tabular data, to estimate a variable value's importance we compute a measure of \emph{evidence gain} from knowing the value, relative to not knowing it. Formally, our importance function is the difference between the function value at the original input and the expected function value for the input with variable $j$ removed. The expectation is taken over a distribution generated by an imputation model conditioned on the remaining covariates.
\begin{align*}
    Imp&ortance(\mathbf{x}_{i,j}) = \nonumber \\  
    &f(\mathbf{x}_i) - \mathbb{E}_{p(\mathbf{x}_{i,j} | \mathbf{x}_{i,-j})} f(\mathbf{x}_{i,-j} \cup \mathbf{x}_{i,j})
\end{align*}
where $p(\mathbf{x}_{i,j}|\mathbf{x}_{i,-j})$ is given by a multinomial logistic regression fit to the training data, and $\mathbf{x}_{i,-j}$ is the data point without feature $j$, and $f(\mathbf{x}_{i,-j} \cup \mathbf{x}_{i,j})$ is the data point $\mathbf{x}_{i,-j}$ with feature value $\mathbf{x}_{i,j}$ imputed at index $j$. We choose to use logistic regressions with no feature engineering in order to 1) generate calibrated probability distributions, and 2) scale straightforwardly with dataset size.

\begin{figure}[t]
    \centering
    \includegraphics[width=0.4\textwidth]{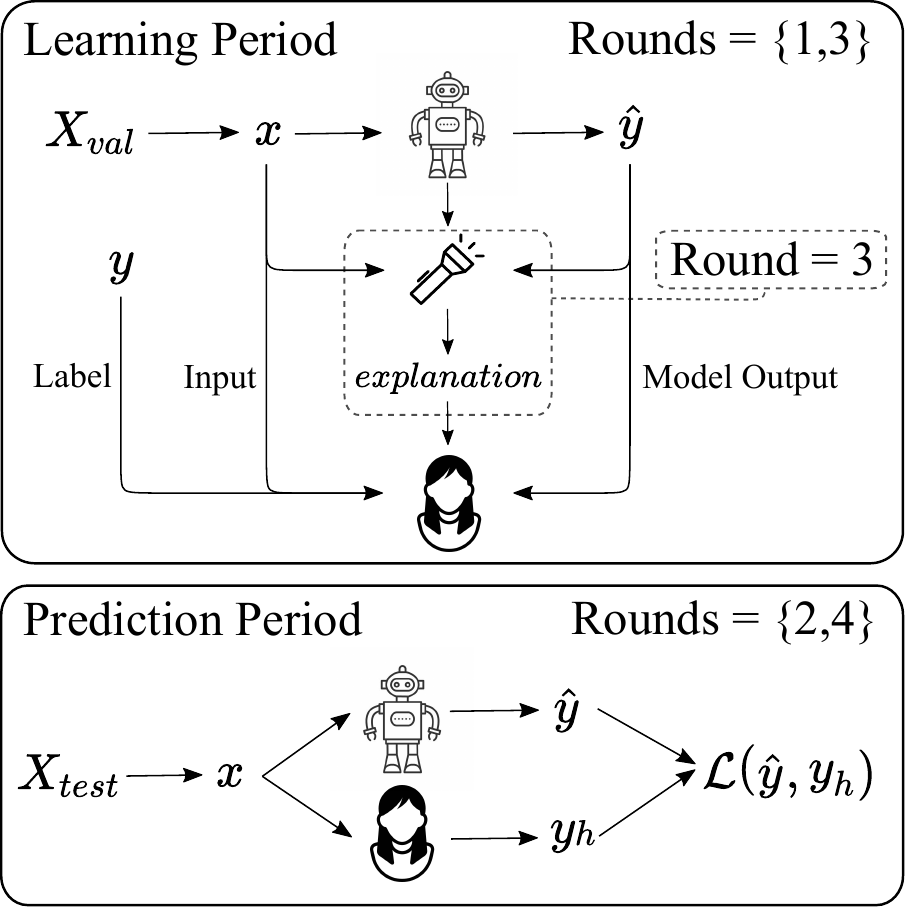}
    \caption{Forward simulation test procedure. We measure human users' ability to predict model behavior. We isolate the effect of explanations by first measuring baseline performance after users are shown examples of model behavior (Rounds 1, 2), and then measuring performance after they are shown \emph{explained} examples of model behavior (Rounds 3, 4).}
    \label{fig:forward}
\end{figure}

\begin{figure}[t]
    \centering
    \includegraphics[width=0.4\textwidth]{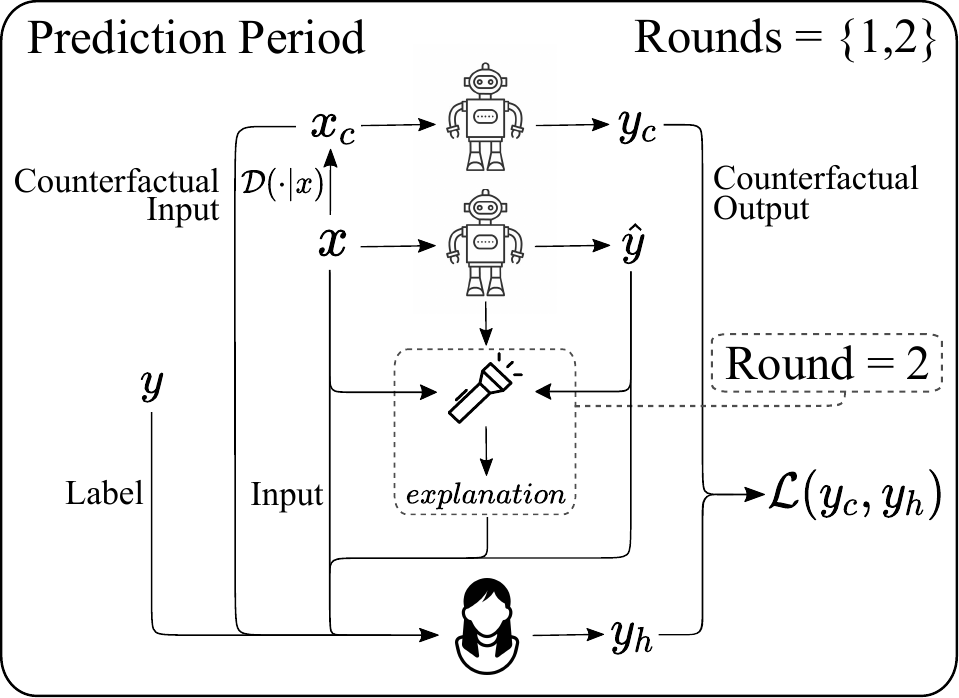}
    \caption{Counterfactual simulation test procedure. Users see model behavior for an input, then they predict model behavior on an edited version of the input. We isolate the effect of explanations by measuring user accuracy with and without explanations.}
    \label{fig:counterfactual}
\end{figure}

\paragraph{Decision Boundary Algorithm.} In detail, the algorithm takes as input a data point $x^*$, the classifier $f$, a perturbation distribution $\mathcal{D}(\cdot|x^*)$, and a measure of distance between inputs $d(x_1,x_2)$. We first sample $\{\tilde{x}\}_{i=1}^{10,000}$ from the perturbation distribution around $x^*$. The eligible perturbations to choose from are those with the opposite prediction from the original: $E = \{\tilde{x}_i | f(\tilde{x}_i) \neq f(x^*)\}$. Then using a distance function $d$, we select a counterfactual input as 
\begin{align*}
    x^{(c)} = \min_{\tilde{x}_i \in  E}  d(x^*, \tilde{x}_i)
\end{align*}
We provide a path from $x^*$ to $x^{(c)}$ by greedily picking the single edit from the remaining edits that least changes the model's evidence margin, which is the difference between positive and negative class scores. Our distance function is the count of different features between inputs, plus the squared Euclidean distance between latent representations. The Euclidean distance is on a scale such that it serves as a tie-breaker:
\begin{align*}
    d(x_1,x_2) = &\sum_j \mathbbm{1}(x_{1j} \neq x_{2j}) \\ &+ ||f(x_1) - f(x_2)||_2^2.    
\end{align*}

\begin{figure*}[t!]
    \centering
    \includegraphics[width=0.95\textwidth]{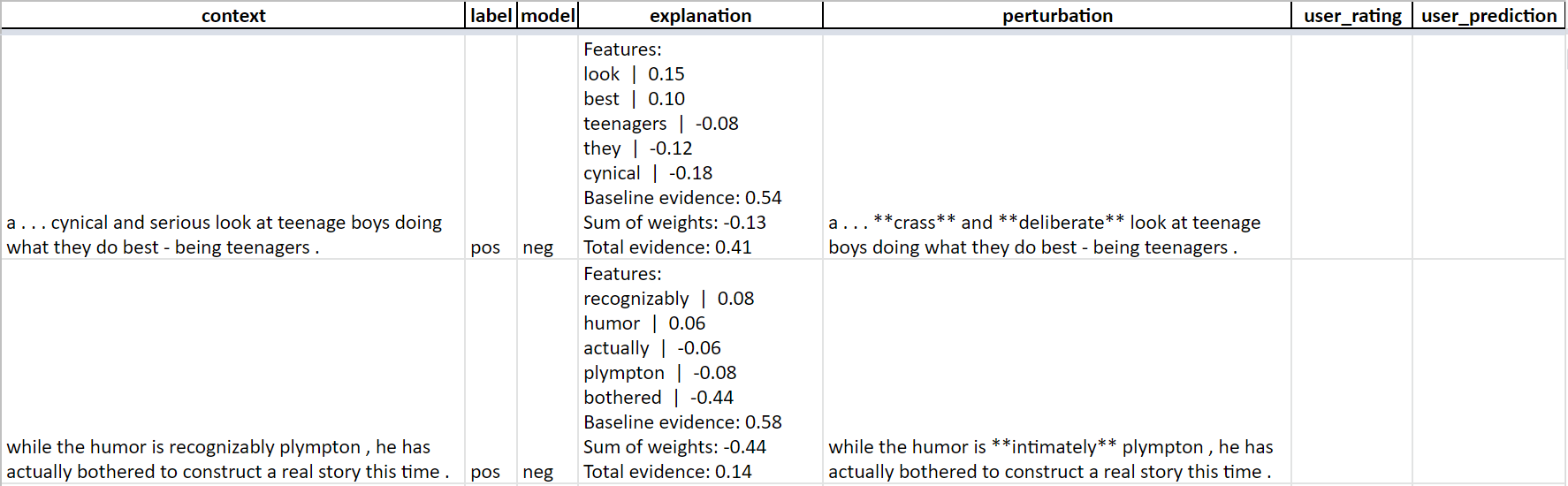}
    \caption{A screenshot of our user testing interface. This example is of the counterfactual Post test with LIME for text data.}
    \label{fig:interface}
\end{figure*}

\subsection{Perturbation Distributions}

\label{sec:perturbation}

We design perturbation distributions for two points in our experiments: 1) selecting counterfactual inputs in simulation tests, and 2) generating decision boundary explanations. First, we describe our approaches for selecting counterfactual inputs, which are conditioned on the need for a certain prediction type: either the same prediction as the original input or the alternative class. In both data domains, we sample $10,000$ local perturbations around the input and then randomly pick a sample that the model predicts to be of the needed prediction type. While working with tabular data, we sample perturbations as follows: we randomly choose to make between 1 and 3 edits, then choose the features to edit uniformly at random, and finally pick new feature values uniformly at random. The only sampling constraint is that a variable cannot be set as its original value. 

For text data, we use a strategy that is similar to sampling from the perturbation distribution in \citet{ribeiro_anchors_2018}, which is to randomly substitute words with their neighbors in GloVe word embedding space, sampling neighbors with probability proportional to their similarity. We make a few changes: we 1) decrease probability of token change with the length of sentence, 2) cap the number of edited words at 5 in the chosen perturbation if possible, and 3) limit edited tokens to be nouns, verbs, adjectives, adverbs, and adpositions. Example perturbations are shown in the example of the user testing interface in Figure \ref{fig:interface}, which is given for a counterfactual test with text data.

\subsection{Simulation Test Design}

In Figures \ref{fig:forward} and \ref{fig:counterfactual}, we include additional representations of our experimental design, showing each test separately and in slightly greater detail than in Figure \ref{fig:procedure}.

\subsection{Testing Environment}

We show a screenshot of our user testing interface in Figure \ref{fig:interface}. This example is of the counterfactual Post test with LIME for text data. Tests are administered through spreadsheets, wherein users read test material and place responses. Users are guided from file to file by the experimenter.

\end{document}